# Learning multi-phase flow and transport in fractured porous media with auto-regressive and recurrent graph neural networks

Mohammed Al Kobaisi[1,2*], Wenjuan Zhang[3], Waleed Diab[1], and Hadi Hajibeygi[4]

[1] Department of Chemical and Petroleum Engineering, Khalifa University of Science and Technology, Abu Dhabi, UAE

[2] Delft Institute of Applied Mathematics, Delft University of Technology, Delft, The Netherlands

[3] School of Civil Engineering, Qingdao University of Technology, Qingdao 266520, China

[4] Department of Geoscience and Engineering, Delft University of Technology, Delft, The Netherlands

* Corresponding Author: M.S.K.AlKobaisi@tudelft.nl; Mohammed.AlKobaisi@ku.ac.ae

## Abstract

In the past three decades, a wide array of computational methodologies and simulation frameworks has emerged to address the complexities of modeling multi-phase flow and transport processes in fractured porous media. The conformal mesh approaches which explicitly align the computational grid with fracture surfaces are considered by many to be the most accurate. However, such methods require excessive fine-scale meshing, rendering them impractical for large or complex fracture networks. The Embedded Discrete Fracture Model (EDFM) framework, on the other hand, offers a good balance between accuracy and efficiency and has gained a lot of traction in recent years. Nonetheless, it is not free of drawbacks; EDFM can, and often will, generate fracture cells that have orders of magnitudes smaller volumes than the matrix cells, which significantly impacts the convergence performance of nonlinear solvers. Such issue is particularly pronounced when discretizing stochastically generated fracture networks, and in uncertainty quantification applications, where a large number of realizations is often required. In this work, we propose to learn the complex multi-phase flow and transport dynamics in fractured porous media with graph neural networks (GNN). GNNs are well suited for this task due to the unstructured topology of the computation grid resulting from EDFM discretization. We propose two deep learning architectures, a GNN and a recurrent GNN. Both networks follow a two-stage training strategy: an autoregressive one step roll-out, followed by a fine-tuning step where the model is supervised using the whole ground-truth sequence. We demonstrate that the two-stage training approach is effective in mitigating error accumulation during autoregressive model rollouts in the testing phase. Our findings indicate that both GNNs generalize well to unseen fracture realizations, with comparable performance in forecasting saturation sequences, and slightly better performance for the recurrent GNN in predicting pressure sequences. While the second stage of training proved to be beneficial for the GNN model, its impact on the recurrent GNN model was less pronounced. Finally, the performance of both GNNs for temporal extrapolation is tested. The recurrent GNN significantly outperformed the GNN in terms of accuracy, thereby underscoring its superior capability in predicting long sequences.

## 1. Introduction

Simulating multi-phase flow and transport in fractured porous media requires sophisticated modeling approaches that can capture the inherent complexities of fluid movement through heterogeneous geological formations. The intricate interplay between fracture networks and the surrounding porous matrix significantly influences fluid distribution and migration; this requires the integration of detailed fracture geometry and connectivity into numerical simulations. Such comprehensive models are vital for predicting reservoir performance and optimizing operational strategies in a range of applications, including sequestration of carbon dioxide in deep saline aquifers, underground hydrogen storage, and geothermal energy extraction. However, the level of detail required in such scenarios is often computationally prohibitive. Consequently, ongoing advancements in simulation techniques of naturally fractured porous media are critical for the efficient management of subsurface resources.

Over the years, numerous conceptual and numerical approaches have been developed for simulating fractured reservoirs. Broadly, these methods can be categorized into three primary groups: equivalent (or effective) medium models, multi-continuum models, and discrete fracture-matrix (DFM) models ; for an in-depth description of these models see the review by Berre et al. [1] and references therein. Equivalent (effective) medium models operate under the assumption that fractures and the surrounding porous matrix are always in local equilibrium. In these models, the distinct properties of fractures are homogenized into a single set of effective parameters that represent the combined behavior of both the fractures and the matrix. [2].

Multi-continuum models, most notably dual-continuum frameworks such as dual-porosity/dual-permeability models, conceptualize the entire system comprising of two, or more, distinct yet interacting continua [3-5]. In these formulations, the fracture network and the surrounding porous matrix are coupled through a transfer function that governs the exchange of mass and energy between the two continua. Multi-continuum models have proved to be valuable for simulating multi-phase flow in fractured reservoirs and are widely employed in the industry. However, their applicability is generally confined to densely fractured reservoirs, as they are less effective in capturing the dominant influences of large, discrete, and highly conductive fractures. Unlike the equivalent or multi-continuum models, DFM models assume complete knowledge of the Discrete Fracture Network (DFN) which is modeled explicitly [6, 7]. The DFM framework offers a more physically realistic representation of fractured reservoirs. However, it is computationally expensive at the field scale, since it requires unstructured meshes that conform to the intricate geometries of fractures. This poses significant challenges in densely fractured reservoirs and often results in an excessive number of mesh cells. To overcome the meshing difficulties in the DFN framework, the Embedded Discrete Fracture-Matrix (EDFM) model [8, 9] and its more recent variant projection-based EDFM (pEDFM) [10] can be employed. These models discretize the porous matrix independently of the fractures and explicitly represent fractures as lower-dimensional surfaces embedded within the simulation grid. Despite the meshing advantages, EDFM models can still produce very small fracture cells with pore volumes several orders of magnitude smaller than those of the matrix cells, particularly when discretizing stochastically generated three-dimensional fracture networks. This disparity can significantly hinder the convergence performance of nonlinear solvers, such as the widely used Newton-Raphson method. Moreover, considering that numerical simulations often require multiple realizations of fracture networks for tasks like uncertainty quantification [11], EDFM models can still be computationally expensive.

Recently, the geoscience community is becoming increasingly focused on developing accurate and efficient surrogate models using deep learning in lieu of computationally expensive numerical simulations. These models are particularly advantageous for applications that require multiple forward simulations, such as data assimilation, uncertainty quantification, and optimization.

There are several classes of deep learning-based surrogate models, most of which rely on the convolutional neural network (CNN) as a core component. The main advantage of CNNs is that they are very efficient learning machines and can provide robust generalization performance for small datasets.

However, CNNs require that the input data be regular image-like structures. When discrete fractures are present and the reservoir model can no longer be treated as regular images, CNN based surrogate models are generally not applicable. Zhu and Zabaras [12] proposed a dense convolutional encoder-decoder deep network as a surrogate model for single phase steady-state flow in 2D. The input to the network is a heterogeneous permeability field and the output is the corresponding pressure and velocity solutions. Both input and output were considered regular images. The model was later extended to multi-phase flow settings where the input is a permeability field plus a time instance, and the output is the corresponding pressure and saturation solutions [13]. Moreover, Zhao et al. [14] proposed image-to-image neural networks to capture the dynamics of $CO_2$ migration in deep (unfractured) saline aquifers.

Another class of deep learning surrogate models are architectures that employ Recurrent Neural Networks (RNNs), and its variants such as the Long Short-Term Memory (LSTM) network. The LSTM is a specialized type of RNN designed to address the problem of vanishing and exploding gradients during long sequence training. To this end, Tang et al. [15] proposed a deep learning-based surrogate model by combining the residual U-Net with the convolutional-LSTM to better capture the temporal dynamics of multi-phase flow problems.

Yet another class of deep learning architecture is neural operators, which is a class of algorithms designed to learn mappings between function spaces and are geared towards scientific computing applications. Neural operators tend to be more data efficient, more accurate, and exhibit great generalization capabilities. To model spatiotemporal dynamics of geological $CO_2$ storage, Wen et al. [16] presented a U-Net enhanced Fourier neural operator (U-FNO), based on the FNO in [17], that has shown excellent performance in modeling many complex physical phenomena. Diab and Al Kobaisi [18] proposed a U-Net enhanced DeepONet [19] architecture, and Jian et al. proposed the Fourier-MIONet architecture [20], both of which were trained using the same dataset as Wen et al., yet showed better accuracy and higher efficiency in terms of training time. Other works that employ neural operators for geological $CO_2$ storage include nested FNO [21], nested Fourier-DeepONet [22], and UU-DeepONet [23] which display strong generalization and temporal extrapolation capabilities. Despite the acclaimed performance of neural operators, they are essentially designed to process structured data on grids, and many of them employ CNN layers.

Unlike traditional neural networks that operate on Euclidean data (e.g., images or text), GNN work on graphs, where nodes represent entities, and edges represent relationships between them. GNN use message passing mechanisms to aggregate and propagate information across nodes, making them well-suited for applications on unstructured grids. GNN were employed by Sanchez-Gonzalez et al. [24] to learn particle simulations where the particles are represented as nodes in a graph. The GNN adopted an Encoder-Processor-Decoder architecture, and the physical dynamics were learned as message passing between nodes and edges of the graph. In a later work, the model was extended to learn from numerical simulations on unstructured meshes [25]. GNN can be combined with other architectures to improve performance in specific use cases; Wu et al. [26] developed a hybrid network for learning subsurface multi-phase flow simulations where a 3D-U-Net is used for learning the pressure solutions and a GNN is used for the saturation solutions on a structured 3D Cartesian grid; Tang and Durlofsky [27] applied a similar GNN to a 2D reservoir discretized by unstructured polygonal meshes; Ju et al. [28] combined a GNN with a graph-based LSTM to learn the spatiotemporal dynamics of $CO_2$ storage in a 2D unstructured mesh constrained by faults.

Given the widespread adoption of EDFM models for simulating flow and transport in fractured reservoirs, there is significant interest in leveraging GNN to learn multi-phase flow dynamics from EDFM simulation data. EDFM models produce a large number of non-neighboring connections which significantly differ from 2D unstructured meshes commonly used in the literature. In this work, we propose to train GNN surrogate models using numerical simulation data from EDFM models.

From a fundamental perspective, and assuming that the boundary conditions and well control parameters are fixed, numerical reservoir simulation data can be viewed as a mapping between an input $x$ and an output $y$, where the input $x$ represents time-invariant geological reservoir properties $m$ and an initial reservoir state $y^0$; and the output $y$ represents the sequence of spatial-temporal evolution of reservoir states (pressure, phase saturation, etc.,) $y = (y^1, ..., y^{n_T})$ where $n_T$ is the total number of time steps. When training a deep learning model, the target is to train the network to approximate the mapping between the input $x$ and output sequence $y$ [16, 18]. However, this approach can be inefficient especially when the length of the output sequence is large. Furthermore, generating the entire output sequence simultaneously disregards the distinct nature of the temporal dimension by treating it equivalently to all other dimensions. This approach inhibits the deep learning model from effectively capturing the underlying temporal dynamics, ultimately degrading its extrapolation performance.

To enhance the ability of deep learning models to capture the temporal dynamics of the problem, a common approach is to account for the sequential nature of the prediction task by training the surrogate model in an autoregressive manner. In this approach, the model takes the reservoir properties and the current reservoir state as inputs and predicts the reservoir state at the next time step. This approach was adopted in [29] and [13] where a surrogate model was trained to take the static reservoir properties (permeability) and the solution state at time instance $t$ as input, and output the reservoir states at the subsequent time step. During inference, the surrogate model operates analogously to conventional numerical simulators, where the predicted output at the current time step serves as the input for the subsequent time step. When a RNN (or one of its variants) is employed in the architecture, the deep learning model maintains an internal memory of historical information and makes predictions one step at a time.

In this work, we employ both the autoregressive strategy and the RNN architecture to train surrogate GNN models and evaluate their performance in predicting the spatiotemporal evolution of reservoir states. The rest of the paper is organized as follows: a brief introduction of the mathematical model is given in Section 2, followed by the details of surrogate GNNs in Section **Error! Reference source not found.**, and model training in Section 4. Experimental results are presented in Section 5, and Section 6 concludes the paper.

## 2. Problem definition

We consider the problem of simulating immiscible, two-phase flow in a fractured reservoir. The phases are denoted by $o$ and $w$, respectively. The mass conservation equation is written for each phase $\alpha = o, w$ as:

$$\frac{\partial}{\partial t}(\phi \rho_\alpha S_\alpha) + \nabla \cdot (\rho_\alpha \mathbf{v}_\alpha) = \rho_\alpha q_\alpha . \tag{1}$$

Here, $\phi$ is porosity, $\rho_\alpha$ is mass density of phase $\alpha$, $S_\alpha$ is phase saturation, $\mathbf{v}_\alpha$ is phase velocity and $q_\alpha$ is phase volumetric source/sink terms. The phase velocity is related to the phase pressure $p_\alpha$ by Darcy's law:

$$\mathbf{v}_\alpha = -\frac{k_{r\alpha}}{\mu_\alpha} \mathbf{K} \nabla p_\alpha , \tag{2}$$

where $\mathbf{K}$ is the absolute permeability tensor, and $k_{r\alpha}$, $\mu_\alpha$ are phase relative permeability and viscosity, respectively. The system is closed by the following constraints:

$$S_w + S_o = 1\,,$$
$$p_o - p_w = p_{cow}(S_w), \tag{3}$$

where $p_{cow}(Sw)$ denotes the capillary pressure between the two phases, along with appropriate boundary and initial conditions.

The mathematical model can be solved numerically by the cell-centered finite volume method based on Two-Point Flux Approximation (TPFA). The reservoir is first partitioned into a set of non-overlapping control volumes and the fracture networks are then modeled by the Embedded Discrete Fracture Model (EDFM). Discrete fracture cells are formed, and matrix-fracture, fracture-fracture connections are introduced as non-neighboring connections in addition to the usual matrix-matrix connections. As a result, the control volumes are connected in an unstructured way in the computational graph. Phase pressure $p_o$ and saturation $S_w$ are usually chosen as the primary unknowns at each control volume. The governing equations can then be discretized by the fully implicit finite volume scheme, leading to a system of nonlinear equations in terms of the primary unknowns that can be solved by some nonlinear solver, e.g., Newton-Raphson, to advance the discrete solutions forward in time. Additional details of the numerical scheme can be found in [30].

## 3. Network architectures

The construction of a graph representation ($G$) of the reservoir to learn the multi-phase flow dynamics using a GNN involves defining node features ($V$), edge features ($E$), and the incident matrix ($A$) representing the connectivity information of the graph. We setup the graph for discrete time step ($n$) which results in the following formulation for the graph:

$$G^n = (V^n, E, A), \tag{4}$$

where $V^n$ is the set of node features at time step $n$, and it is obtained by stacking the node features $\mathbf{v}_i^n$ for all the nodes. The node feature vector $\mathbf{v}_i^n$ for any node $i$ is given by:

$$\mathbf{v}_i^n = [v_i, \phi_i, \log_{10}(k_i), \mathbf{n}_i, y_i^n]\,. \tag{5}$$

Here, $v_i, \phi_i$ and $k_i$ are the bulk volume, porosity and permeability of cell $i$, respectively. $\mathbf{n}_i \in \mathbb{R}^3$ is a one-hot encoding vector to differentiate cells containing source terms, cells containing sink terms and cells containing neither. Separate GNNs are built to learn the pressure and saturation solutions, and $y_i^n$ denotes either pressure or saturation of cell $i$ at time step $n$, depending on the GNN. Similarly, the set of edge features $E$ is obtained by stacking the edge feature vector $\mathbf{e}_{ij}$ for all the edges. For an edge connecting node $i$ and $j$, the edge feature vector $\mathbf{e}_{ij}$ is given by:

$$\mathbf{e}_{ij} = \left[\log_{10}(T_{ij}), \mathbf{p}_j - \mathbf{p}_i, \|\mathbf{p}_j - \mathbf{p}_i\|_2\right], \tag{6}$$

where $T_{ij}$ is the static transmissibility of the connection, and $\mathbf{p}_{i,j}$ denotes the position vector of the centroids of cell $i$ and cell $j$. Note that the edge features ($E$) are independent of time, hence the absence of the superscript $n$. We denote the reservoir properties that are independent from the pressure and saturation by $m$. In this setup, each computational cell becomes a vertex in the graph and two directed edges are formed if two cells are connected either directly in the mesh or indirectly by non-neighboring connections. The graph $G^n$ can be obtained directly from $m$ and $y^n$ using equations **Error! Reference source not found.** and **Error! Reference source not found.** and written as a mapping:

$$(m, y^n) \rightarrow G^n = (V^n, E, A).$$ (7)

## 3.1 Graph neural network

Training a neural network autoregressively means that predictions at the current time step depend on past predictions, such that the output updates sequentially over time. If we denote the GNN by $g_\theta$ where $\theta$ is the network parameters, the graph representation at each time step $n$ by $G^n$, and the decoder by $d_\phi$ where $\phi$ are the decoder's parameters, we can state the input and the output of the network as:

$$\Delta y_i^{n+1} = d_\phi \big( g_\theta(G^n) \big),$$
$$\hat{y}^{n+1} = \Delta y_i^{n+1} + y^n,$$ (8)

where $\hat{y}^{n+1}$ denotes the predicted reservoir state at the time step $n+1$. Note here that the decoder $d_\phi$ translates the embeddings of the GNN ($g_\theta$) into changes in the reservoir state (e.g., how pressure and saturation evolve over time). The overall mapping between the input and output can be obtained by combining equations (7) and (8); see Figure 1 for a graphical illustration. If we denote this mapping by $f_{\mathrm{ar}}$, we can then write the entire model as follows:

$$\hat{y}^{n+1} = f_{\mathrm{ar}}(m, y^n).$$ (9)

Given a reservoir realization $m$ and an initial reservoir state $y^0$, the GNN can be used autoregressively to predict the reservoir states at future time steps in the following way:

$$\begin{cases} \hat{y}^0 = y^0 \\ \hat{y}^n = f_{\mathrm{ar}}(m, \hat{y}^{n-1}), n = 1, 2, \dots, n_T \end{cases},$$ (10)

where $n_T$ is the total number of time steps.

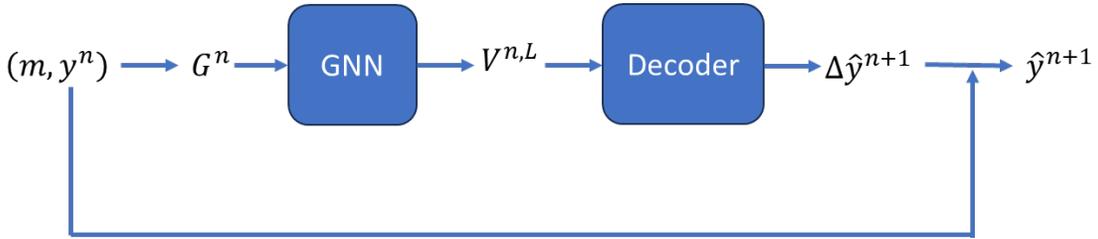

**Figure 1:** Overall structure of the autoregressive GNN.

Our GNN adopts the Encoder-Processor architecture in [25] and is shown in Figure 2. The encoder consists of node and edge encoders that transform their respective features into latent embeddings. These embeddings pass through multiple processor blocks, which update node and edge representations simultaneously. Finally, the output node embeddings from the last processor block go through a decoder to produce the final output of the graph neural network.

The node and edge encoders are implemented using multilayer perceptron (MLP):

$$\mathbf{v}_i^{n,0} = \mathrm{MLP}_v^0(\mathbf{v}_i^n),$$ (11)

$$\mathbf{e}_{ij}^0 = \mathrm{MLP}_e^0 \big( \mathbf{e}_{ij} \big),$$

where $\mathbf{v}_i^{n,0}$ and $\mathbf{e}_{ij}^0$ are encoded node and edge embeddings, respectively. The processor consists of $L$ identical blocks, each with its own trainable parameters. For the $l^{th}$ processor block ($l = 1, \ldots, L$), node and edge embeddings are updated using two MLPs with residual connections as follows:

$$\mathbf{e}_{ij}^l = \mathrm{MLP}_e^l \big( \big[ \mathbf{e}_{ij}^{l-1}, \mathbf{v}_i^{n,l-1}, \mathbf{v}_j^{n,l-1} \big] \big) + \mathbf{e}_{ij}^{l-1},$$

$$\mathbf{v}_i^{n,l} = \mathrm{MLP}_v^l \left( \left[ \mathbf{v}_i^{n,l-1}, \sum_{j \in \mathcal{N}_i} \mathbf{e}_{ij}^l \right] \right) + \mathbf{v}_i^{n,l-1}, \tag{12}$$

where $\mathbf{e}_{ij}^l$ and $\mathbf{v}_i^{n,l}$ are the output embeddings of the $l^{th}$ processor block, $\mathcal{N}_i$ denotes the set of neighboring nodes to node $i$, and $[\cdot]$ denotes concatenation in the feature dimension. $\mathrm{MLP}_e^l$ and $\mathrm{MLP}_v^l$ denote the MLP for updating the edge and node embeddings for the $l^{th}$ processor block, respectively. The final node embeddings of the last processor block are decoded into the solution space using a MLP:

$$\Delta y_i^{n+1} = \mathrm{MLP}_v^{L+1} \big( \mathbf{v}_i^{n,L} \big). \tag{13}$$

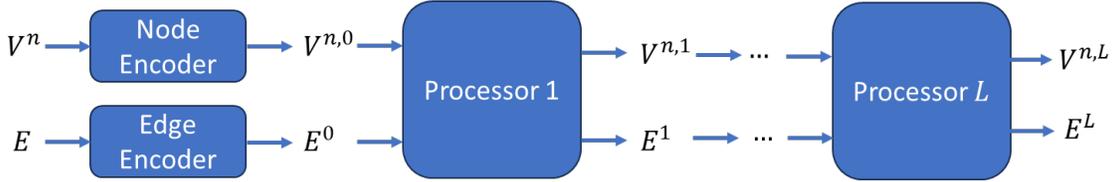

**Figure 2:** Structure of the graph neural network.

### 3.2 Recurrent graph neural network

Autoregressive training is known to be prone to error accumulation; this is because the predictions of neural networks will always contain some errors, and as the model is rolled out these errors tend to accumulate. Noise injection [25] is one strategy to control error accumulation and stabilize the model rollout. Another strategy involves adopting a recurrent neural network architecture to maintain a memory of past rolled out states which would limit error accumulation. Integrating a GNN with a recurrent structure effectively reduces error accumulation and improves stability when rolled out over multiple future steps, as shown in [28]. Therefore, we adopt a recurrent GNN in this work to evaluate its effectiveness. We combine a long short-term memory (LSTM) recurrent network with the GNN introduced in section 3.1.

The LSTM architecture contains a structured gating mechanism. The gate mechanism consists of a forget gate ($f$), which discards irrelevant information, an input gate ($i$), which regulates the addition of new information, and an output gate ($o$), which determines the contribution of the updated cell state to the hidden state $H$. The cell state $C$ serves as a memory unit, which allows the network to retain essential information over long time sequences. At each time step $n$, the graph $G^n$ is processed by the GNN module to produce the node embeddings $V^{n,L}$, which act as input to the LSTM block to obtain the updated cell state $C^n$ and hidden states $H^n$. Finally, the hidden states $H^n$ pass through a decoder to get the change in reservoir states. The LSTM block consists of layers of LSTM cells and the formulation of the $l^{th}$ layer of a LSTM cell can be written as:

$$
\begin{aligned}
i &= \sigma\big(W_{xi}X^n + W_{hi}H^{n-1,l} + b_i\big), \\
f &= \sigma\big(W_{xf}X^n + W_{hf}H^{n-1,l} + b_f\big), \\
o &= \sigma\big(W_{xo}X^n + W_{ho}H^{n-1,l} + b_o\big), \\
C^{n,l} &= f \odot C^{n-1,l} + i \odot \tanh\big(W_{xg}X^n + W_{hg}H^{n-1,l} + b_g\big), \\
H^{n,l} &= o \odot \tanh\big(C^{n,l}\big),
\end{aligned}
\tag{14}
$$

where $\sigma$ is the sigmoid activation function, and $\odot$ denotes the Hadamard product. The input $X^n$ is equal to $V^{n,L}$ for $l = 1$ and $H^{n,l-1}$ otherwise. The output of the last layer of the LSTM cell is the final output $C^n$ and $H^n$ of the LSTM block.

The authors in [28] stacked a GNN on top of a graph convolutional recurrent LSTM block proposed in [31]. The LSTM block uses the Chebyshev spectral graph convolution kernel to process the input data. However, the Chebyshev spectral graph convolution requires the computation of the eigenvalues of the graph Laplacian, which can be quite expensive for large graphs. Considering the fact that the message passing GNN has already extracted node embeddings utilizing spatial information and that the recurrent network is mainly used to learn temporal evolution of the reservoir states at the node level, we built the recurrent GNN in a similar way to "Model 1" of the graph convolutional recurrent network in [31]. Hence, the LSTM block only contains linear layers and the computation of eigenvalues of graph Laplacian is thus avoided. The overall structure of the model is shown in Figure 3. The overall mapping of the recurrent GNN can be written as:

$$
\hat{y}^{n+1} = f_{\text{lstm}}(m, y^n, C^n, H^n).
\tag{15}
$$

The rollout of the recurrent GNN is like equation (10) and can be written as:

$$
\begin{cases}
\hat{y}^0 = y^0, \\
\hat{y}^n = f_{\text{lstm}}(m, \hat{y}^{n-1}, C^{n-1}, H^{n-1}), n = 1,2,\dots,n_T,
\end{cases}
\tag{16}
$$

where the cell state $C^0$ and hidden state $H^0$ of the LSTM block are initialized to zero matrices.

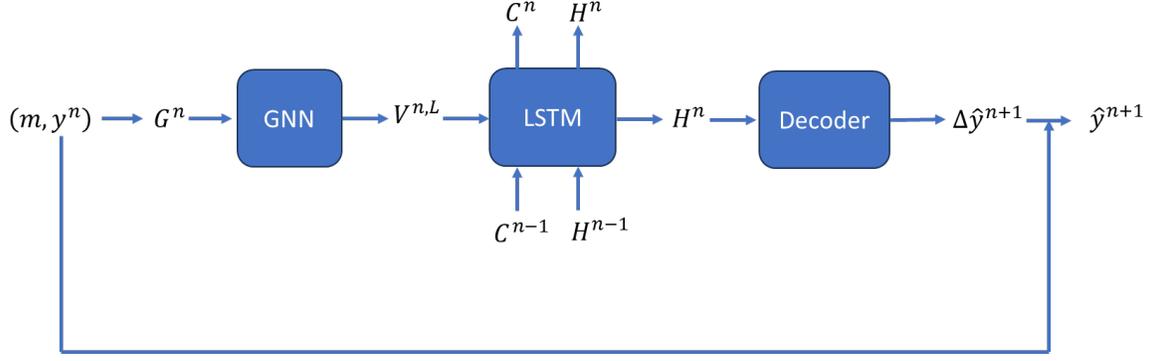

**Figure 3:** Structure of the recurrent graph neural network.

## 4. Model training

Both the standalone GNN and the GNN with LSTM are trained autoregressively. Given a reservoir realization and an initial state $y^0$, both models would predict the next time step in the sequence $y = (y^1, ..., y^{n_T})$. As discussed earlier, this training methodology is prone to errors because each new rolled out state $(y^n + \epsilon)$ is dependent on a previously rolled out state $(y^{n-1} + \epsilon)$ that the model predicted. Since each rolled out state inherently contains some small error $(\epsilon)$, every new rolled out state amplifies this error further leading to high levels of error propagation into future time steps. To combat error accumulation, we propose a two-stage training procedure. In the first stage, the model is trained with ground truth states at every time step instead of its own predictions. This is known as teacher forcing for the training of recurrent neural networks. We adopt the teacher forcing training strategy to avoid introducing artificial noise which magnitude is not easily determined a priori and to achieve faster convergence. In the second stage, we fine tune the GNN by rolling out the entire sequence of predictions autoregressively, i.e., the model is supervised using the whole sequence. Specifically, the loss functions for the two stages of training can be written as:

$$
\begin{aligned}
\mathcal{L}_{\text{stage1}} &= \frac{1}{N n_T} \sum_{i=1}^{N} \sum_{j=1}^{n_T} l\big(f_{\text{gnn}}(m^i, y^{i,j-1}), y^{i,j}\big) \ , \\
\mathcal{L}_{\text{stage2}} &= \frac{1}{N n_T} \sum_{i=1}^{N} \sum_{j=1}^{n_T} l\big(\hat{y}^{i,j}, y^{i,j}\big) ,
\end{aligned}
\tag{17}
$$

where $N$ is the number of reservoir realizations and $n_T$ is the total number of time steps which is the same as the length of the target sequence. $f_{\text{gnn}}$ denotes the GNN and the GNN with LSTM. $y^{i,j}$ and $\hat{y}^{i,j}$ are the ground truth and predicted reservoir state at time step $j$ for the $i^{th}$ realization, respectively. Note that $\hat{y}^{i,j}$ is obtained by rolling the respective GNNs autoregressively from the initial state $y^0$. The loss term $l$ for a generic prediction $\hat{y}$ and its label $y$ is given by:

$$
l(\hat{y}, y) = \frac{1}{n_c} \|\hat{y} - y\|_2^2 + \frac{1}{n_c} \|\hat{y} - y\|_1 \ ,
\tag{18}
$$

where $\|\cdot\|_2$ and $\|\cdot\|_1$ are the $L^2$ and $L^1$ norms, respectively, and $n_c$ is the dimension of $y$ and $\hat{y}$. Similar to [23], we also found that adding mean absolute error to the mean squared error can improve the performance of the neural network.

## 5.  Experimental results

### 5.1  Data generation

The open-source MATLAB Reservoir Simulation Toolbox (MRST) [30] was used to generate the training data for the graph neural networks developed in this work. The reservoir is box-shaped with physical dimensions of $500m \times 500m \times 5m$ and discretized by a $50 \times 50 \times 1$ Cartesian grid. The rock matrix has a constant permeability of $k_m = 50$md and a constant porosity of $\phi = 0.25$. The discrete fracture network was generated stochastically using the open-source MATLAB code ADFNE [32]. The fracture planes are general polygons in 3D space and two sets of orthogonal discrete fractures are generated. **Error! Reference source not found.** shows the top view of several realizations of the discrete fracture networks. The fractures appear to be line segments in the figure since all the fracture planes are nearly vertical. All the fracture planes have an aperture of 1 mm, a porosity of 0.8 and a permeability of $10^7$md. The fluid properties are given in **Error! Reference source not found.**. The initial reservoir pressure is 10 MPa and the initial water saturation is 0.2. A rate-controlled injector is placed at the lower left corner and a bottomhole pressure-controlled producer at the upper right corner. The injection rate is constant and a total of 0.5 pore volume of water is injected over the entire simulation period. The bottom hole pressure of the producer is fixed at 8 MPa. The total simulation time is 5 years discretized into 30 equal time steps.

Figure **5** shows the production rate over time for a subsample of realizations of the fracture networks. The production rate increases at first and then plateaus before dropping again, but the actual values vary significantly for different fracture network realizations. Figure 6 shows the pressure and saturation solutions at four different time instances corresponding to time steps $n = 1, 5, 10,$ and 20 for one model realization. As depicted in the figure, the saturation field is highly irregular because of the presence of the discrete fracture network. We ran the simulations for 500 different fracture network realizations and stored the pressure and saturation solutions at each time step. Pressure and saturation values from time step 0 to 20 are then processed for training and testing our GNNs, rendering a sequence of length 21. The first 10 time-steps (corresponding to 11 reservoir states in total) are used during the training. Moreover, out of the 500 simulation runs, 400 runs are used for training, 50 runs for validation, and the remaining 50 runs are reserved for testing.

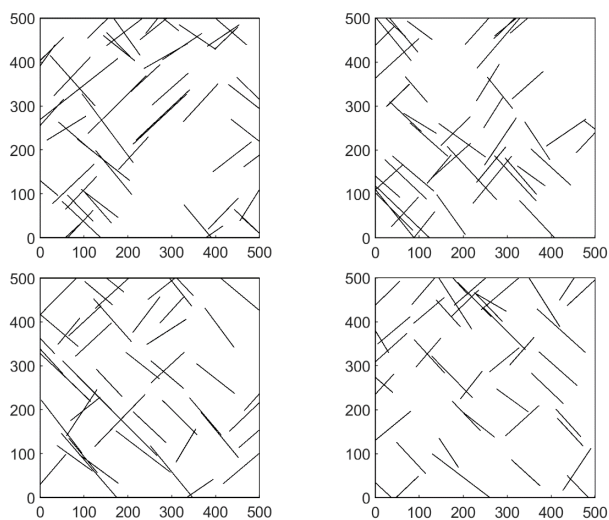

**Figure 4:** Realizations of discrete fracture networks.

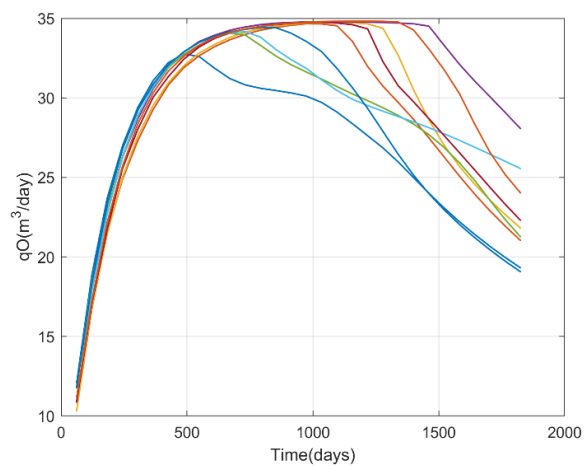

**Figure 5:** Production profile for different fracture network realizations

Table 1: Fluid properties

| Property | Value |
|---|---|
| Water viscosity | 1 cp |
| Oil viscosity | 5 cp |
| Water surface density | 1000 kg/m$^3$ |
| Oil surface density | 850 kg/m$^3$ |
| Water compressibility | 3e-6 1/psi |
| Oil compressibility | 1e-5 1/psi |
| Relative permeability exponent | 2 |
| Connate water saturation | 0.2 |
| Residual oil saturation | 0.2 |

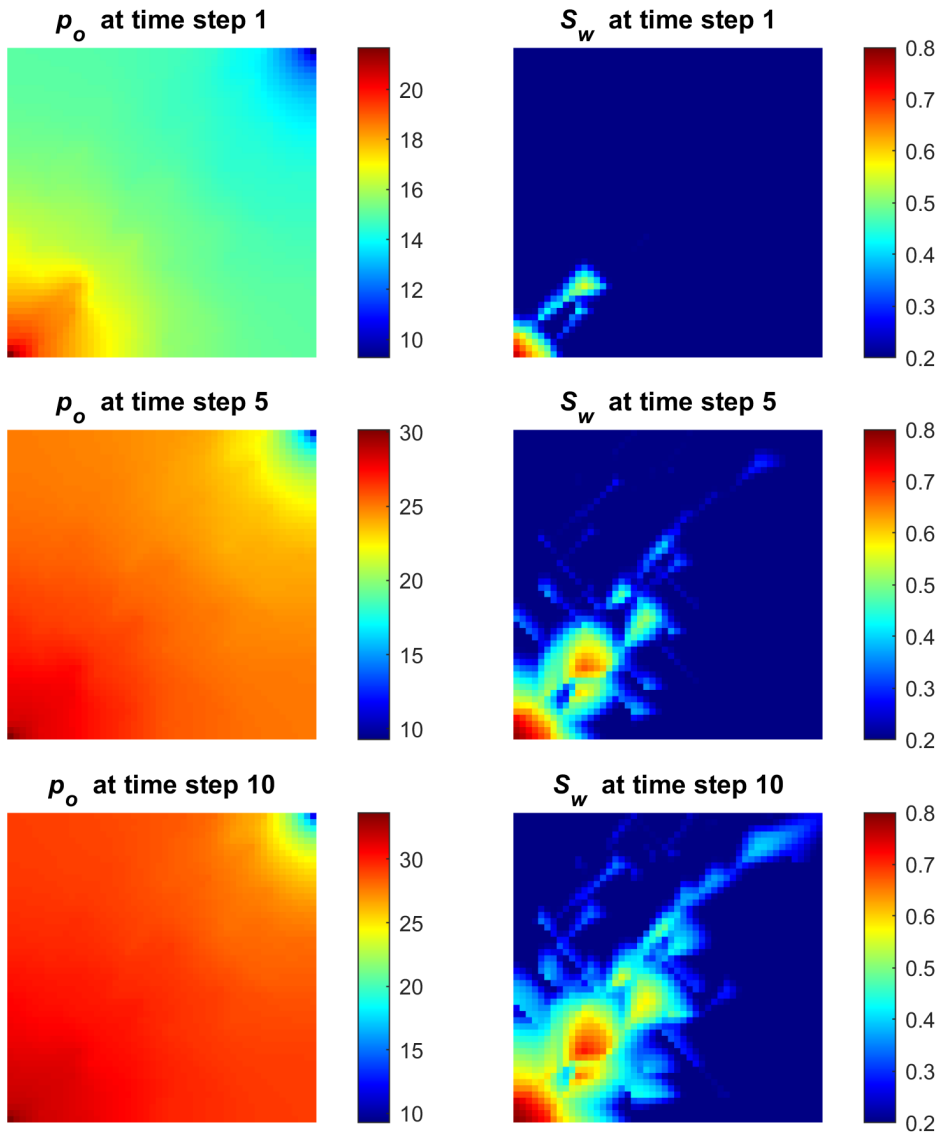

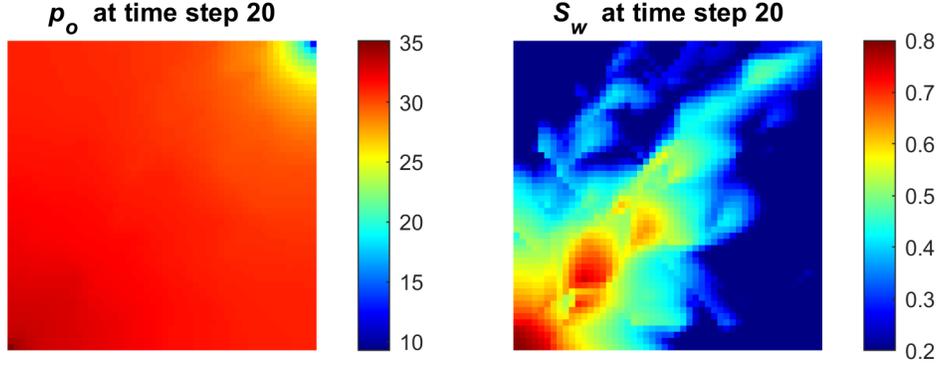

**Figure 6:** Pressure and saturation solutions at time step 1, 5, 10 and 20 for one model realization.

### 5.2 Model setup

The GNN consists of a GNN module and a decoder while the recurrent GNN is composed of a GNN moule, a decoder and a LSTM block. Both models share an identical architecture for the GNN module and the decoder. The only difference is the additional LSTM block for the recurrent GNN. The trainable parameters of the GNN module and the decoder are associated with the various MLPs used. **Error! Reference source not found.** lists the details of the MLPs' structures used in each component of the GNN module and the decoder. The input sizes for the node and edge encoders are determined by the dimension of the node feature and edge feature, respectively. The hidden size and output size are the same for all the MLPs except for the output of the decoder. The modeling of the pressure sequences and the saturation sequences are separated completely. The hidden dimension $h$ is taken as 40 for pressure and 48 for saturation, and the GNN module consists of 12 processors for pressure and 8 processors for saturation. The LSTM block in the recurrent GNN is composed of 2 layers of LSTM cell. The total number of trainable parameters in the GNN is 188,201 for pressure and 184,753 for saturation. In comparison, the total number of trainable parameters in the recurrent GNN is 214,441 for pressure and 222,385 for saturation. The GNNs were implemented using the open-source python library Pytorch and Pytorch Geometric. The trainable parameters were optimized using the Adam optimizer with the scheduler *ExponentialLR*. For the first stage of one-step training, the optimizer has a learning rate of 1e-3 with a weight decay of 5e-3. The batch size is 4 and the models are trained for 200 epochs. The loss term is computed every 5 epochs on the validation dataset, and we save the model that has the smallest validation loss to avoid overfitting. For the second stage of training, the models are trained for 100 epochs, and the learning rate is decreased to 1e-4 while the weight decay remained at 5e-3. The models were trained on a NVIDIA Tesla V100 GPU. The total training time in hours for the various GNN models are listed in Table 3.

Table 2: MLP architecture hyperparameters

| MLP | Input size | Hidden size | Output size | Number of layers | Activation function |
|---|---|---|---|---|---|
| **Node encoder** | 7 | h | h | 3 | relu |
| **Edge encoder** | 5 | h | h | 3 | relu |
| **Node processor** | 2h | h | h | 3 | relu |
| **Edge processor** | 3h | h | h | 3 | relu |
| **decoder** | h | h | 1 | 3 | relu |

**Table 3:** Training time in hours for the GNNs

| | *GNN* | | *Recurrent GNN* | |
|---|---|---|---|---|
| | Stage 1 (hours) | Stage 2 (hours) | Stage 1 (hours) | Stage 2 (hours) |
| $p_o$ | 4.30 | 2.15 | 3.30 | 1.72 |
| $S_w$ | 3.40 | 1.70 | 2.69 | 1.41 |

To evaluate the performance of the neural networks, the mean relative absolute error (MRAE) is used as the accuracy metric, and it is given by:

$$e_p^n = \frac{1}{n_c} \sum_{i=1}^{n_c} \frac{|\hat{p}_{o,i}^n - p_{o,i}^n|}{p_{o,i}^n},$$
$$e_s^n = \frac{1}{n_c} \sum_{i=1}^{n_c} \frac{|\hat{S}_{w,i}^n - S_{w,i}^n|}{S_{w,i}^n}, \qquad (19)$$

where $n_c$ is the total number of cells. $p_{o,i}^n, S_{w,i}^n$ are the ground truth pressure and saturation solutions of cell $i$ at time step $n$, respectively, and the corresponding network predictions are denoted by $\hat{p}_{o,i}^n$ and $\hat{S}_{w,i}^n$.

### 5.3   Results and discussions

We test the performance of both GNNs for two tasks. The first task is generalization which we define as a model's ability to predict state variables given a new DFN for the same time window it was trained on. The second task is simultaneous extrapolation and generalization which we define as a model's ability to predict state variables given a new DFN for a time window beyond what it was trained on. In both tasks the testing dataset which consists of 50 realizations is used.

To test the generalization performance of the network models for new DFN realizations, the models are rolled out for the first 10 time-steps over the 50 testing trajectories. Box plots of error metrics per rolled out time step for pressure and saturation are shown in Figure 7 andFigure **8**, respectively. The box plots show the distribution of the error metrics computed using equation (19) for 50 testing trajectories as a function of the first 10 time-steps. As expected, the error metrics generally increase over time due to error accumulation with successive model roll outs. This is especially true of the saturation (Figure 8) for both the GNN and the recurrent GNN. Table 4 lists the same error metrics averaged over all 10 time-steps for both models after the two training stages. The results show that the GNN benefits from the second stage of training. However, the recurrent GNN shows less improvement from the second stage of training. Comparing the performance of the autoregressive GNN with that of the recurrent GNN reveals that the latter has better accuracy for modeling the pressure sequence, and the two different types of GNN have comparable performance for the saturation sequence. Figure 9 andFigure **10** show the testing results of the two GNNs with two stages of training for a specific DFN realization.

**Table 4:** Error metrics averaged over 10 time steps for the generalization test

|  | GNN | | Recurrent GNN | |
|---|---|---|---|---|
|  | Stage 1 | Stage 2 | Stage 1 | Stage 2 |
| $p_o$ | 0.076 | 0.052 | 0.044 | 0.044 |
| $S_w$ | 0.058 | 0.050 | 0.059 | 0.053 |

To test the performance of the GNNs in the simultaneous extrapolation and generalization task, we use the trained models to predict the last ten pressure and saturation states in the sequence of the testing dataset. For the GNN, only the pressure/saturation ground truth states at the tenth time step needs to be provided as input and the model is then rolled out autoregressively to predict future states. For the recurrent GNN models, the ground truth values of the first eleven reservoir states (corresponding to the first ten time-steps) are provided as input to update the memory states of the LSTM block first, and the models are then rolled out autoregressively. Box plots of error metrics per extrapolated time step for pressure and saturation are shown in Figure 11 and Figure **12**, respectively. Table 5 shows the error metrics averaged over extrapolated time steps for both models. The results in the table indicate that the second stage of training does not improve the accuracy of the models in the simultaneous extrapolation and generalization task. This may be caused by the fact that stage 2 training focuses on the entirety of the first ten time-steps, rendering the models less capable of extrapolating into future time steps.

Considering stage one training only, the recurrent GNN model clearly outperforms the GNN in the simultaneous extrapolation and generalization task. Figure 11 and Figure **12** show that the errors for both pressure and saturation stay within acceptable ranges for the recurrent GNN, demonstrating its superior performance in predicting future reservoir states that are beyond the time frame of training compared to the autoregressive models. The results suggest that the memory states of the LSTM block derived from the first ten reservoir states in the sequence is very beneficial for predicting future reservoir states. To visually inspect the quality of the solution, we plot the pressure/saturation maps and the corresponding prediction error maps of the GNN models for a particular DFN realization in Figure 13 and Figure **14**. A comparison of the recurrent graph neural network (GNN) performance in the generalization task (Table 4) and the simultaneous extrapolation and generalization task (Table 5) reveals that the recurrent GNN excels in extrapolation. This superior performance can be attributed to the long short-term memory (LSTM) component of the network, which enables it to retain and leverage flow patterns from previous time steps effectively.

**Table 5:** Error metrics averaged over 10 time steps for the extrapolation test

|  | GNN | | Recurrent GNN | |
|---|---|---|---|---|
|  | Stage 1 | Stage 2 | Stage 1 | Stage 2 |
| $p_o$ | 0.061 | 0.091 | 0.019 | 0.016 |
| $S_w$ | 0.058 | 0.231 | 0.033 | 0.066 |

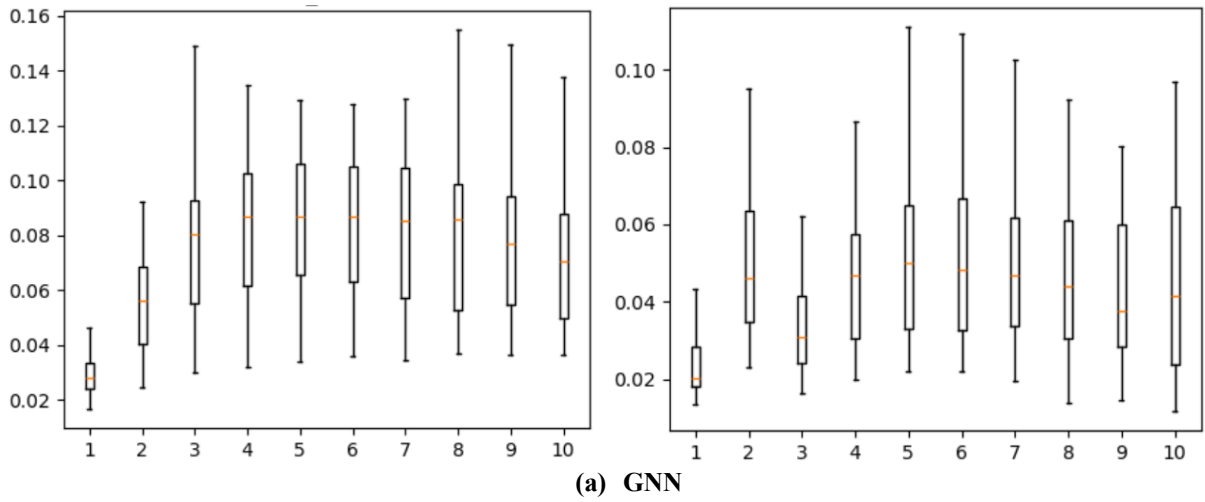

**(a) GNN**

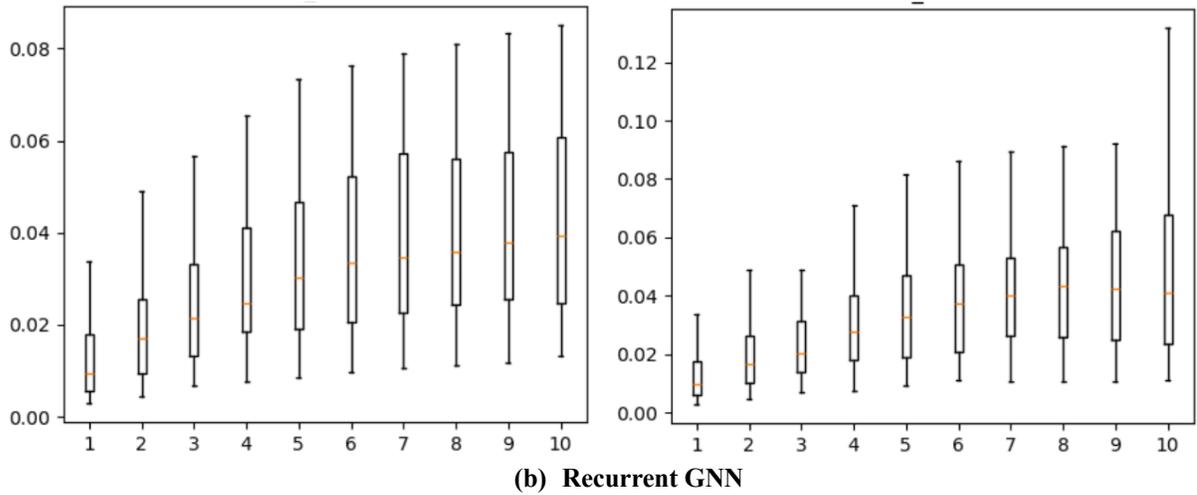

**(b) Recurrent GNN**

**Figure 7:** Box plots of pressure error metrics for (a) GNN and (b) recurrent GNN. The models are rolled out 10 steps for 50 testing trajectories. The left column corresponds to models after stage 1 training and the right column stage 2 training.

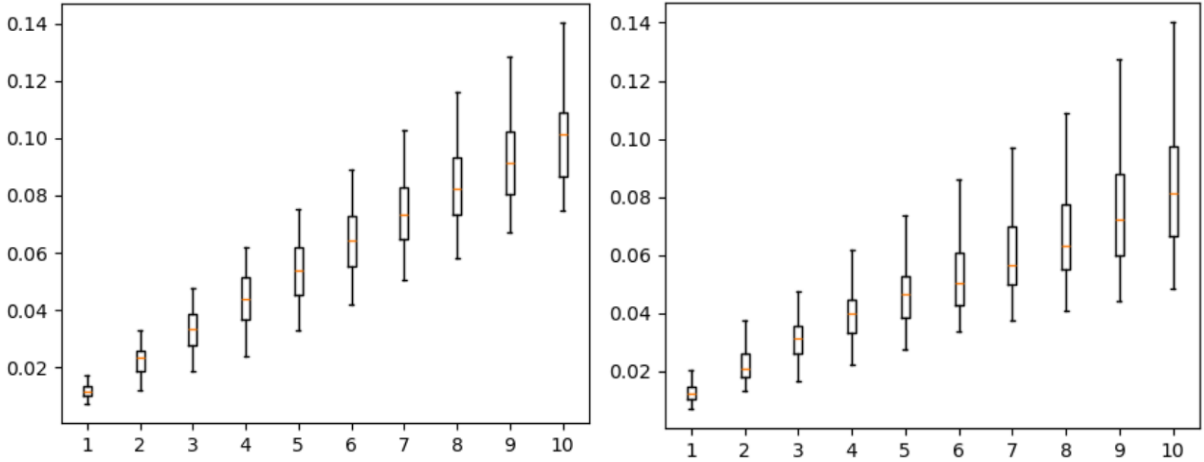

**(a) GNN**

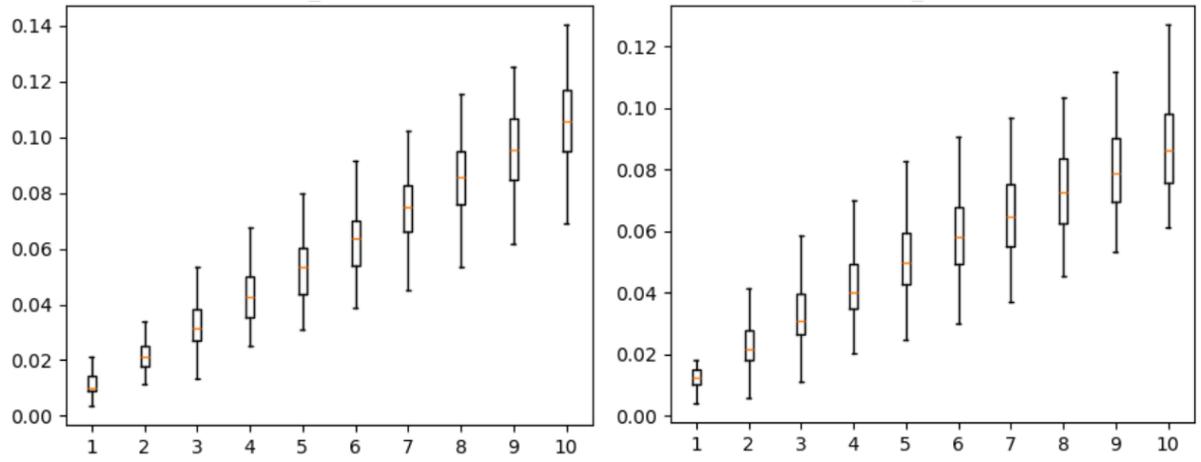

**(b) Recurrent GNN**

**Figure 8:** Box plots of saturation error metrics for GNN (a) and recurrent GNN (b). The models are rolled out 10 steps for 50 testing trajectories. The left column corresponds to models after stage 1 training and the right column after stage 2 training.

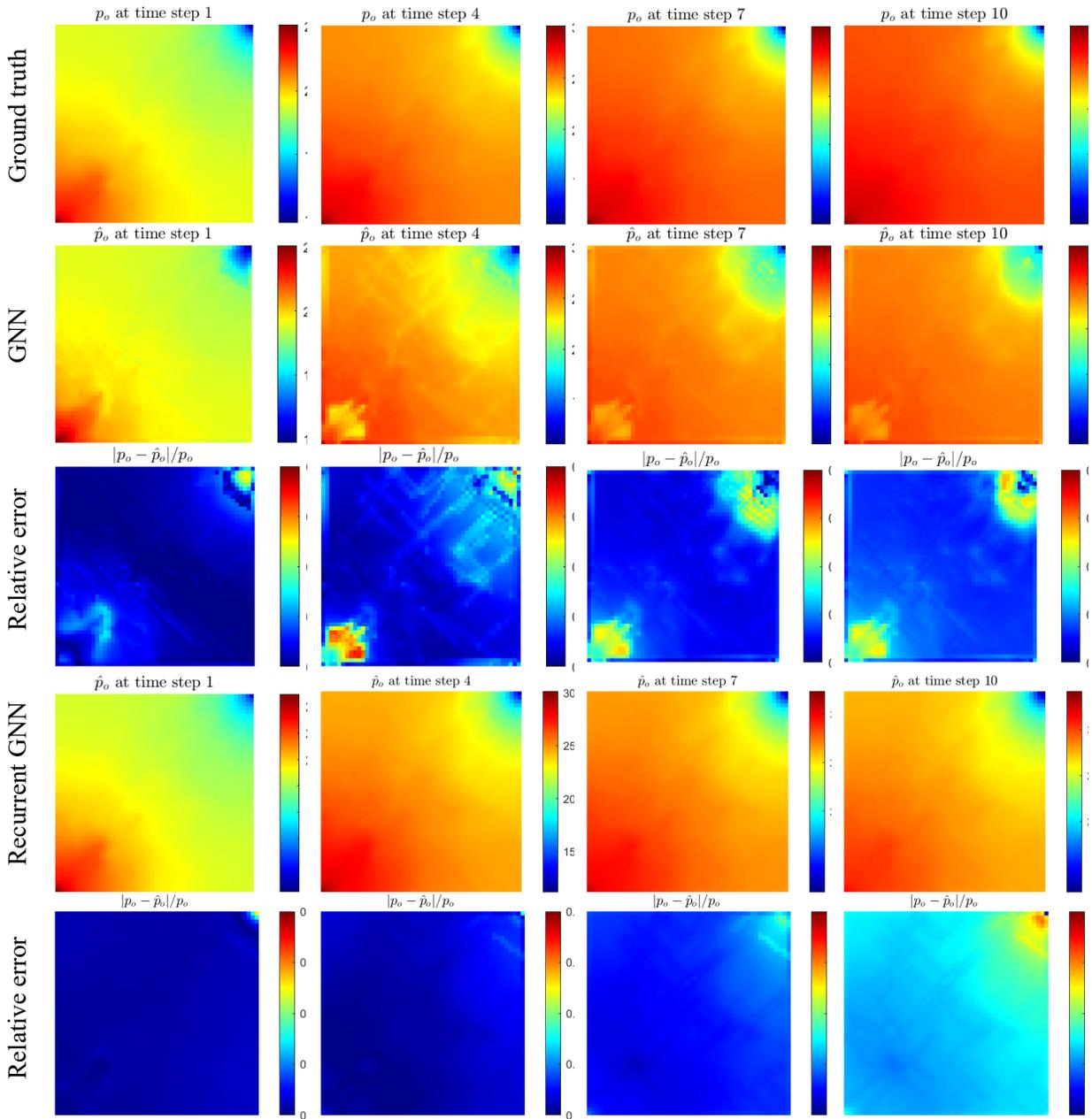

**Figure 9:** Pressure predictions of GNN models for one DFN realization at time steps 1, 4, 7 and 10. From first to fifth row: ground truth of pressure map; prediction of GNN; relative error of GNN; prediction of recurrent GNN; relative error of recurrent GNN.

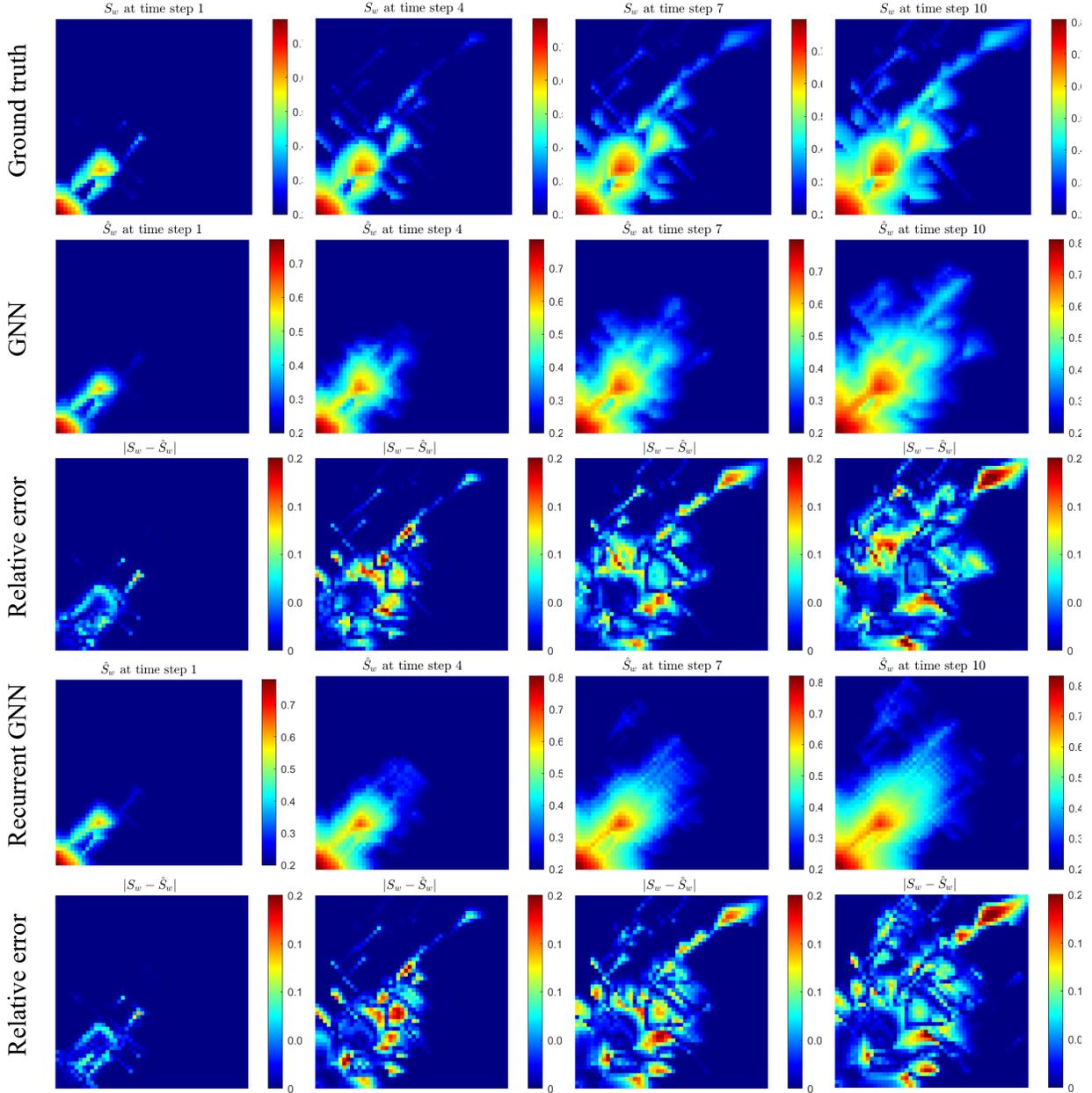

**Figure 10:** Saturation predictions of GNN models for one DFN realization at time step 1, 4, 7 and 10. From first to fifth row: ground truth of saturation map; prediction of GNN; absolute error of GNN model; prediction of recurrent GNN; absolute error of recurrent GNN.

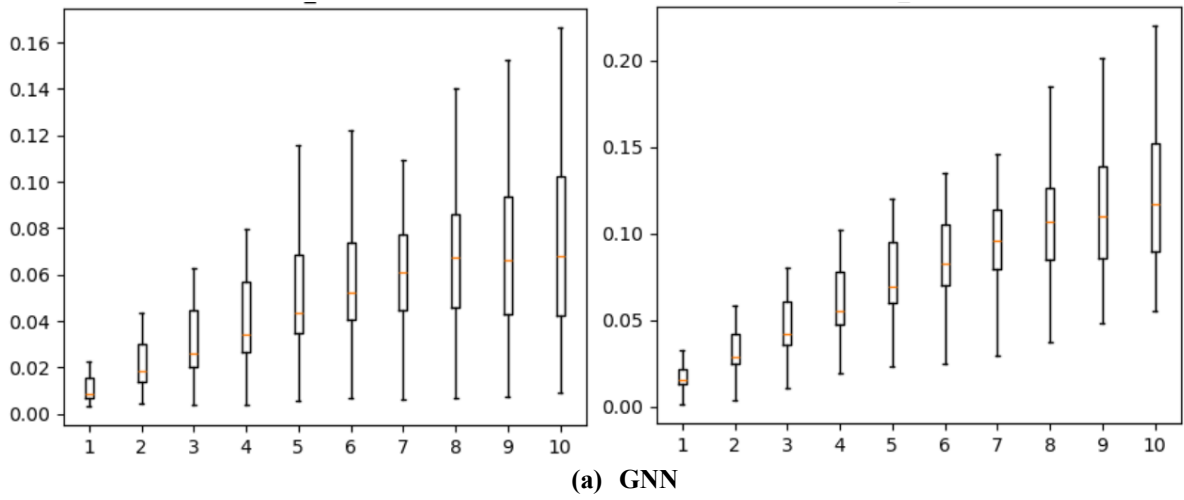

**(a) GNN**

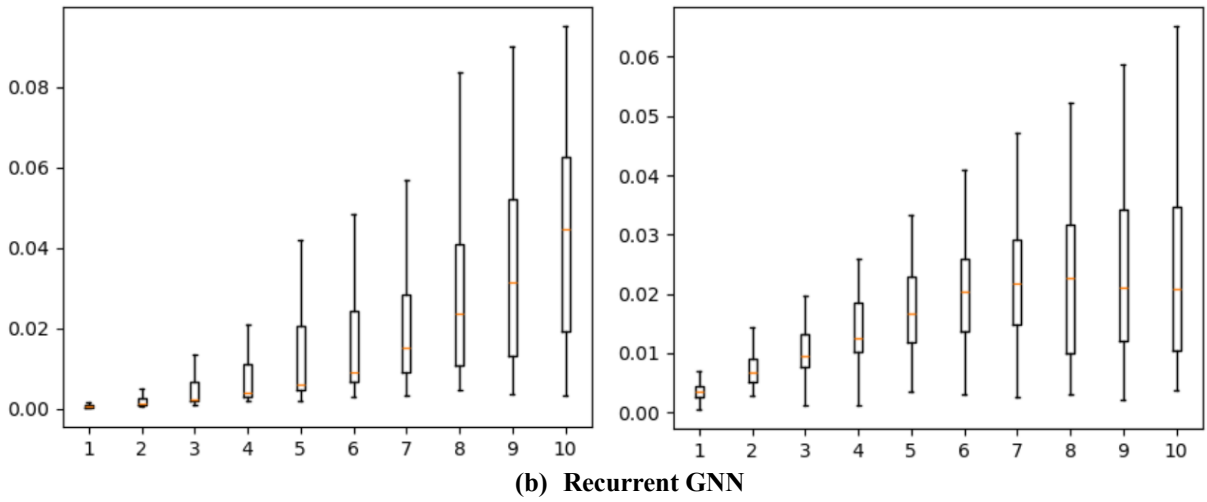

**(b) Recurrent GNN**

**Figure 11:** Box plots of pressure error metrics for (a) GNN and (b) recurrent GNN. Both networks are extrapolated for 10 time steps for the 50 testing trajectories. The left column corresponds to networks after stage 1 training and the right column stage 2 training.

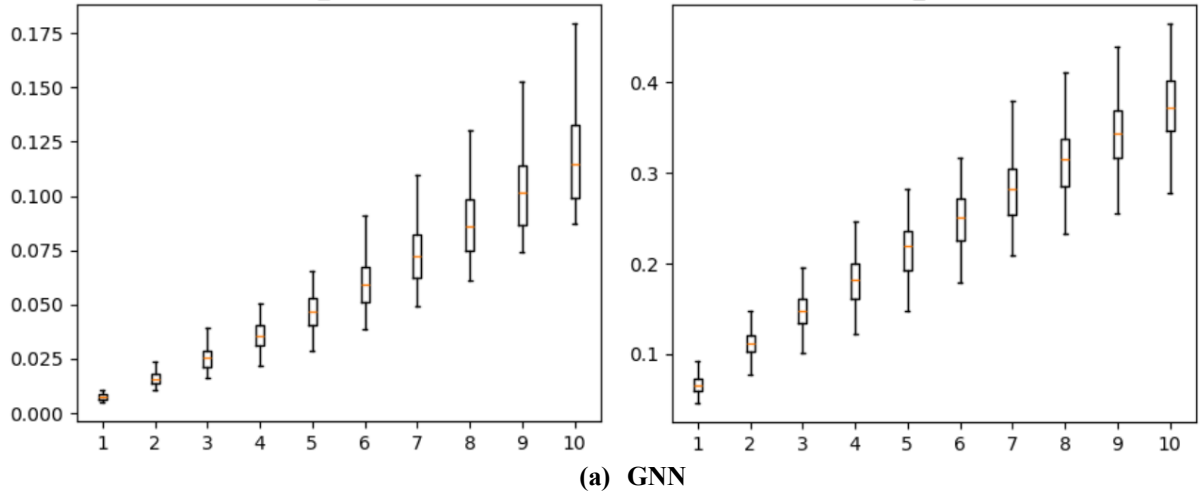

**(a) GNN**

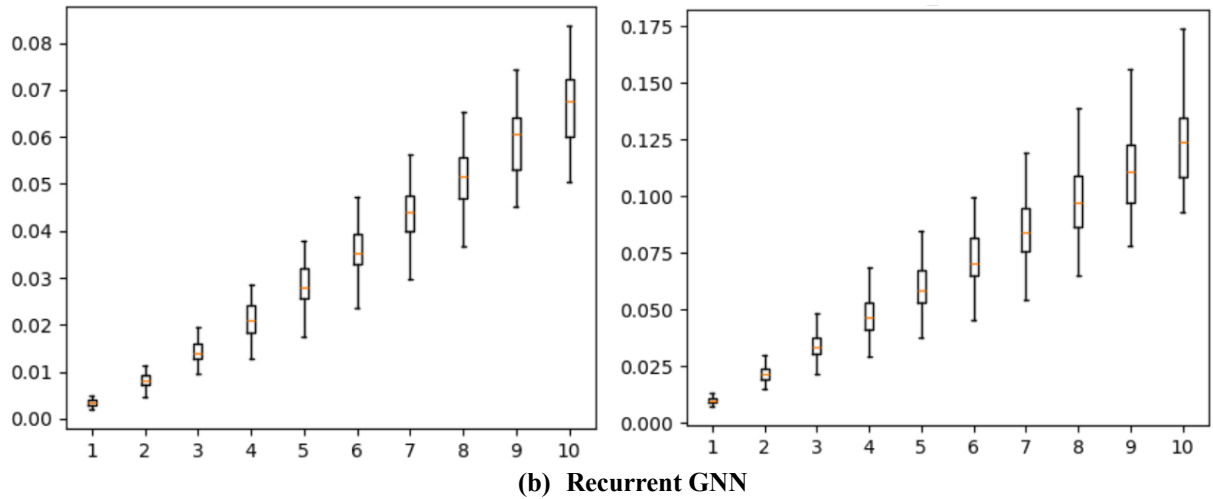

**(b) Recurrent GNN**

**Figure 12:** Box plots of saturation error metrics for (a) GNN and (b) recurrent GNN. Both networks are extrapolated for 10 steps for 50 testing trajectories. The left column corresponds to networks after stage 1 training and the right column stage 2 training.

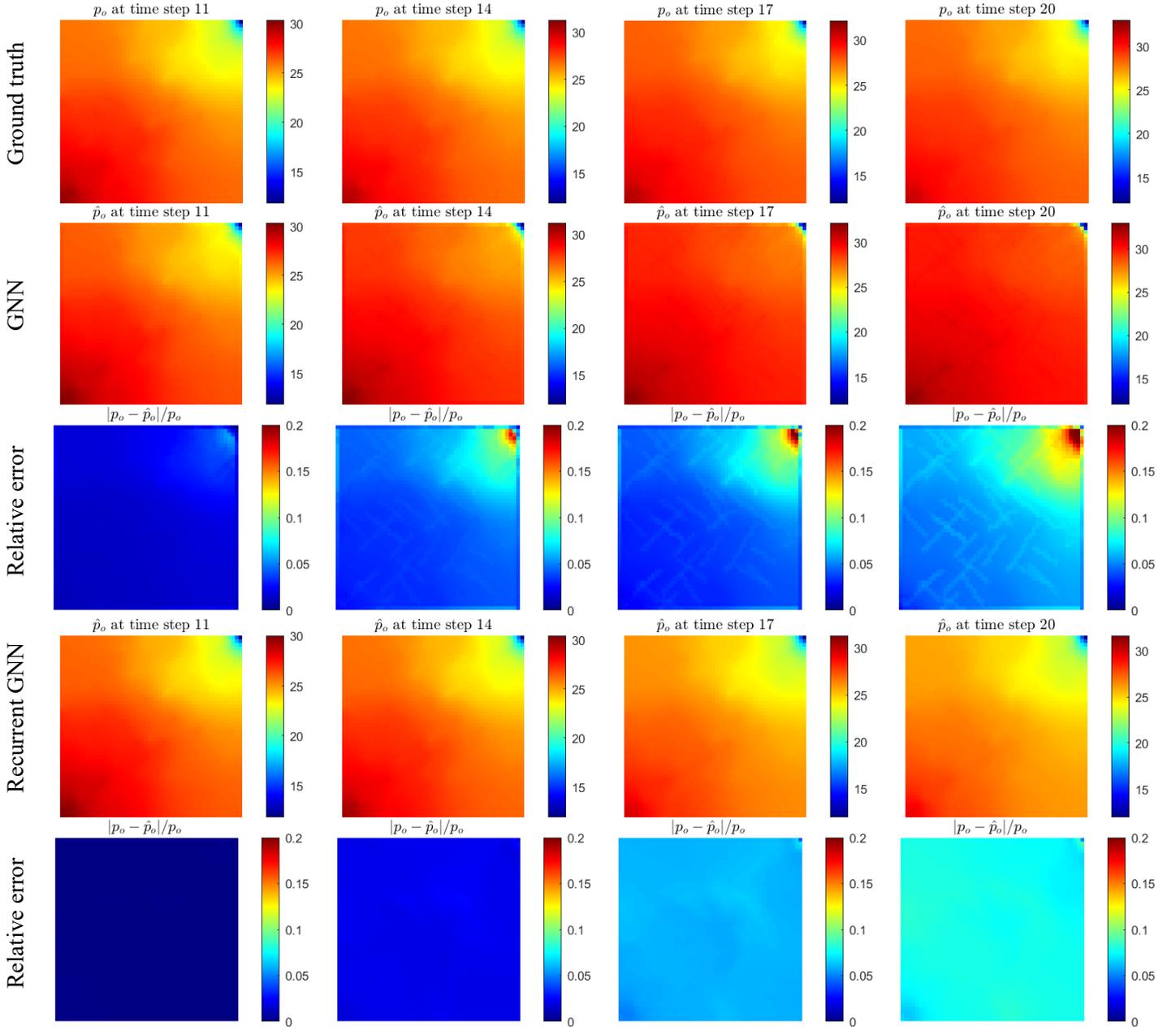

**Figure 13:** Pressure predictions of GNN models for one DFN realization at time step 11,14,17 and 20. From first to fifth row: ground truth of pressure map; prediction of GNN; relative error of GNN; prediction of recurrent GNN; relative error of recurrent GNN.

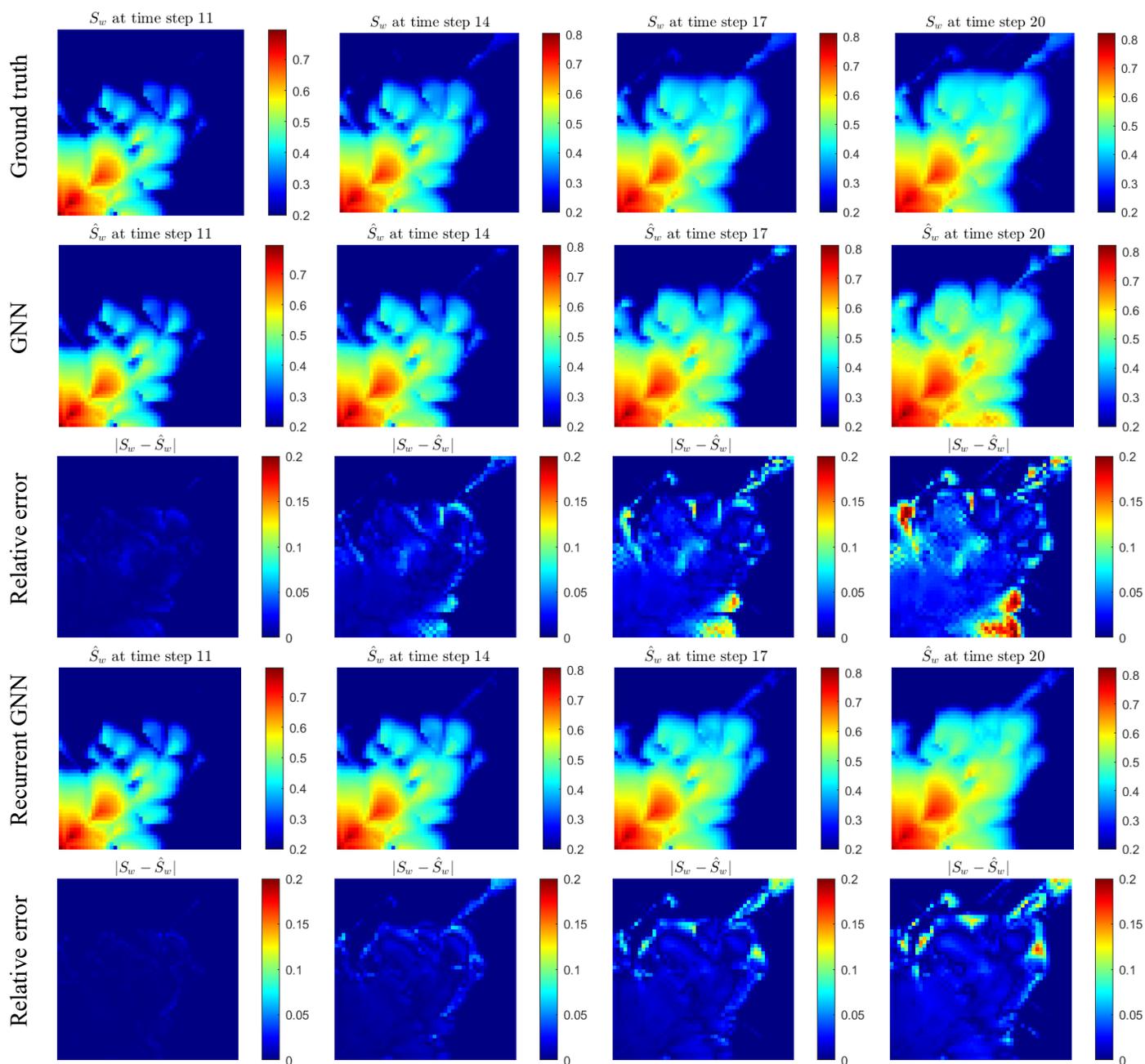

**Figure 14:** Saturation predictions of GNN models for one DFN realization at time step 11,14,17 and 20. From first to fifth row: ground truth of saturation map; prediction of GNN; absolute error of GNN; prediction of recurrent GNN; absolute error of recurrent GNN.

## 6. Conclusions

A GNN and a recurrent GNN are used in this work to learn the pressure and saturation sequences from EDFM multi-phase flow simulation data. Both networks share the same GNN module and decoder; the recurrent GNN also incorporates an LSTM block. Furthermore, a two-stage strategy is devised to train the networks. After training, the GNN are tested for generalization to unseen fracture realizations and for simultaneous generalization and temporal extrapolation performance. The results show that the second stage of training is beneficial to the GNN for generalization but not so much for temporal extrapolation. The effects of the second stage of training on the recurrent GNN are negligible. When used for generalization, the recurrent GNN model has slightly better performance in predicting pressure sequences over the GNN without the recurrent component, and both networks perform comparably in predicting saturation sequences. When used for temporal extrapolation, the recurrent GNN demonstrates a superior performance in predicting both the pressure and the saturation sequences over the GNN due to the LSTM memory which is updated with the reservoir history of all previous time steps. In principle, the prediction of reservoir states at the next time step only requires information of reservoir states at the current time which is the same principle in classical reservoir simulation. In practice, however, carrying historical information by adding a LSTM block seems to help the neural network learn better from data and hence, the recurrent GNN is recommended for the sort of problems considered in this work.

**Acknowledgement**

The authors gratefully acknowledge the computing time made available to them on the Ankabut HPC cluster of Khalifa University of Science and Technology. This work was partly supported by ADNOC (Grant No. 8434000476), the National Natural Science Foundation of China (Grant No. 52304030) and the Natural Science Foundation of Shandong Province, China (Grant No. ZR2023QA034).